\def\year{2022}\relax
\documentclass[letterpaper]{article} 
\usepackage{aaai22}  
\usepackage{times}  
\usepackage{helvet}  
\usepackage{courier}  
\usepackage[hyphens]{url}  
\usepackage{graphicx} 
\usepackage{natbib}  
\usepackage{caption} 
\usepackage{color} 
\usepackage{multirow}
\usepackage{amsmath}
\DeclareCaptionStyle{ruled}{labelfont=normalfont,labelsep=colon,strut=off} 
\usepackage{algorithm}
\usepackage{algorithmic}
\usepackage[T1]{fontenc}

\newcommand{\Add}[1]{\textcolor{black}{#1}}

\usepackage{newfloat}
\usepackage{listings}
\floatstyle{ruled}
\newfloat{listing}{tb}{lst}{}
\floatname{listing}{Listing}
\title{ReProHRL: Towards Multi-Goal Navigation in the Real World using \\ Hierarchical Agents }
\author {
    Tejaswini Manjunath,
    Mozhgan Navardi,
    Prakhar Dixit,
    Bharat Prakash,
    Tinoosh Mohsenin
}
\affiliations {
    University of Maryland, Baltimore County \\
    tejaswini@umbc.edu, mnavardi1@umbc.edu, pdixit1@umbc.edu,
    bhp1@umbc.edu, tinoosh@umbc.edu
}

\usepackage{bibentry}

\begin{document}

\maketitle

\begin{abstract}
Robots have been successfully used to perform tasks with high precision. In real-world environments with sparse rewards and multiple goals, learning is still a major challenge and Reinforcement Learning~(RL) algorithms fail to learn good policies. Training in simulation environments and then fine-tuning in the real world is a common approach. However, adapting to the real-world setting is a challenge. In this paper, we present a method named \textbf{Re}ady for \textbf{Pro}duction \textbf{H}ierarchical \textbf{RL}~(\textbf{ReProHRL}) that divides tasks with hierarchical multi-goal navigation guided by reinforcement learning. We also use object detectors as a pre-processing step to learn multi-goal navigation and transfer it to the real world. Empirical results show that the proposed ReProHRL method outperforms the state-of-the-art baseline in simulation and real-world environments in terms of both training time and performance. Although both methods achieve a 100\% success rate in a simple environment for single goal-based navigation, in a more complex environment and multi-goal setting, the proposed method outperforms the baseline by 18\% and 5\%, respectively. For the real-world implementation and proof of concept demonstration, we deploy the proposed method on a nano-drone named Crazyflie with a front camera to perform multi-goal navigation experiments.

\end{abstract}

\section{Introduction}

Reinforcement Learning~(RL) is a popular method used for a variety of applications like robotics, autonomous navigation, games, energy management, and so on. Autonomous drone navigation is one such widely used application that enables many capabilities like search and rescue~\cite{zuluaga2018deep},  gas leakage detection~\cite{duisterhof2021sniffy}, etc. Despite achieving great success in such applications, training the agent is a compute-intensive and time-consuming process because of the large amount of trial and error required to learn new policies. Recent works in Sim2Real~\cite{truong2021bi, zhang2021sim2real, zhang2020sim2real} have shown promising results in learning navigation and manipulation policies where the Reinforcement learning based policy is trained in simulators and then transferred to real-world environments. However, such Sim2Real adaptation is still not trivial due to the lack of high-fidelity replication of real-world scenes and the imprecise simulation models.

With progress in computer vision research, Unmanned Aerial Vehicles~(UAVs) with computer vision capabilities have gained a lot of interest~\cite{duisterhof2021tiny, navardi2022optimization}. UAVs equipped with high-resolution cameras can be used in real-world environments for a number of applications like transportation systems, search and rescue, agricultural and land surveys, safety, etc. Recent proposals such as You Only Look Once (YOLO)~\cite{benjumea2021yolo} have achieved efficient and fast object detection by obtaining cost-free region proposals sharing full-image convolutional features with the detection network. One of the advantages of using YOLO-based object detection is the availability of pre-trained models trained on millions of real-world images. Only minimal fine-tuning is necessary to fit the model to different use cases.

Training in simulation and deploying in the real world, usually called Sim2Real~\cite{doersch2019sim2real}, is a popular way to deploy such agents in the real-world and production environments. A lot of existing work in image-based robotic navigation tends to use raw images  to learn policies, whether they are in the simulation or the real world. But, the visual gap between the simulation and the real world makes Sim2Real deployment challenging. Therefore, it is important to address this challenge to have a Ready for Production~(RePro) agent. Another challenge is that real-world tasks are complex and have longer horizons. But high-level tasks can be broken down into smaller sub-tasks by leveraging the natural hierarchy that exists in most cases. Hierarchical Reinforcement Learning~(HRL) agents are used to decompose complex tasks into multiple sub-tasks~\cite{prakash2021interactive, li2020hrl4in, staroverov2020real, prakash2021semantic, prakashhierarchical}. The agent consists of a high-level policy that provides a sub-goal and a low-level policy that takes the sub-goal and performs primitive actions in the environment.

In this paper, we propose a framework named ReProHRL that trains Ready for Production Hierarchical Reinforcement Learning agents in simulation which can be easily deployed in the real world. The ReProHRL agent has a fixed high-level module that provides sub-goals and a low-level policy that performs those sub-goals. To address the challenges of learning with high-dimensional observations like images, we use an object detection model, YOLO~\cite{benjumea2021yolo}, to pre-process the raw images. We extract object representation in the form of the Bounding box~(Bbox) coordinates to improve generalization and for robust sim-to-real transfer. Few works~\cite{taghibakhshi2021local, nguyen2019reinforcement} have explored this idea, but none have successfully deployed it in a real-world environment for UAVs in the multi-goal setting. Moreover, we show that our approach can be generalized to multiple goals. We one-hot encode the goal vector to represent the goals (skills) and their relations.

One of the crucial requirements for goal-based navigation is obstacle avoidance. While other methods exist that use an image, depth map, or bounding box,  we use a lidar sensor(available in the drone) as a laser ranger.

The main contributions of this paper can be summarized as follows:

\begin{itemize}
    \item A generalized framework to train hierarchical agents in simulation and deploy in the real world for a variety of multi-goal tasks.
    \item Using YOLOv5 as an object detector to pre-process images and greatly reduce the state space to RL. 
    \item A low-level policy that is trained to reach goals using the pre-processed inputs and distance sensors. Additionally, a controller that switches sub-goals to enable sequential goal completion in virtual and physical environments.
    \item A multi-goal simulation environment using Airsim and unreal engine which enables RL research for deployment on UAVs. 
    \item Deployment and evaluation of the proposed technique on a drone named Crazyflie for multi-goal navigation in the real-world environment. 

\end{itemize}

To evaluate the proposed approach, we measured the training success rate of the HRL agent by applying the ReProHRL approach in simple and complex environments with multiple objects. Moreover, we compare the proposed approach with state-of-the-art works~\cite{navardi2022toward, shiri2022efficient}. The experimental results show that the ReProHRL approach reaches a 100\% success rate in different environments. We also deploy the proposed approach in a real-world environment on a drone named Crazyflie~\cite{giernacki2017crazyflie}.  

The rest of the paper is organized as follows:  First, we discuss related work, followed by our proposed approach in Section \textit{Proposed Methodology}.
 We show the experimental setup in detail in Section \textit{Experimental Setup}. Further in Section \textit{Experimental Results}, we discuss our findings based on experiments before we summarize the proposed work in Section \textit{Conclusion}.

\begin{figure}[t]
\centering
\includegraphics[width=1.2\columnwidth, trim = 6cm 11cm 9cm 5cm]{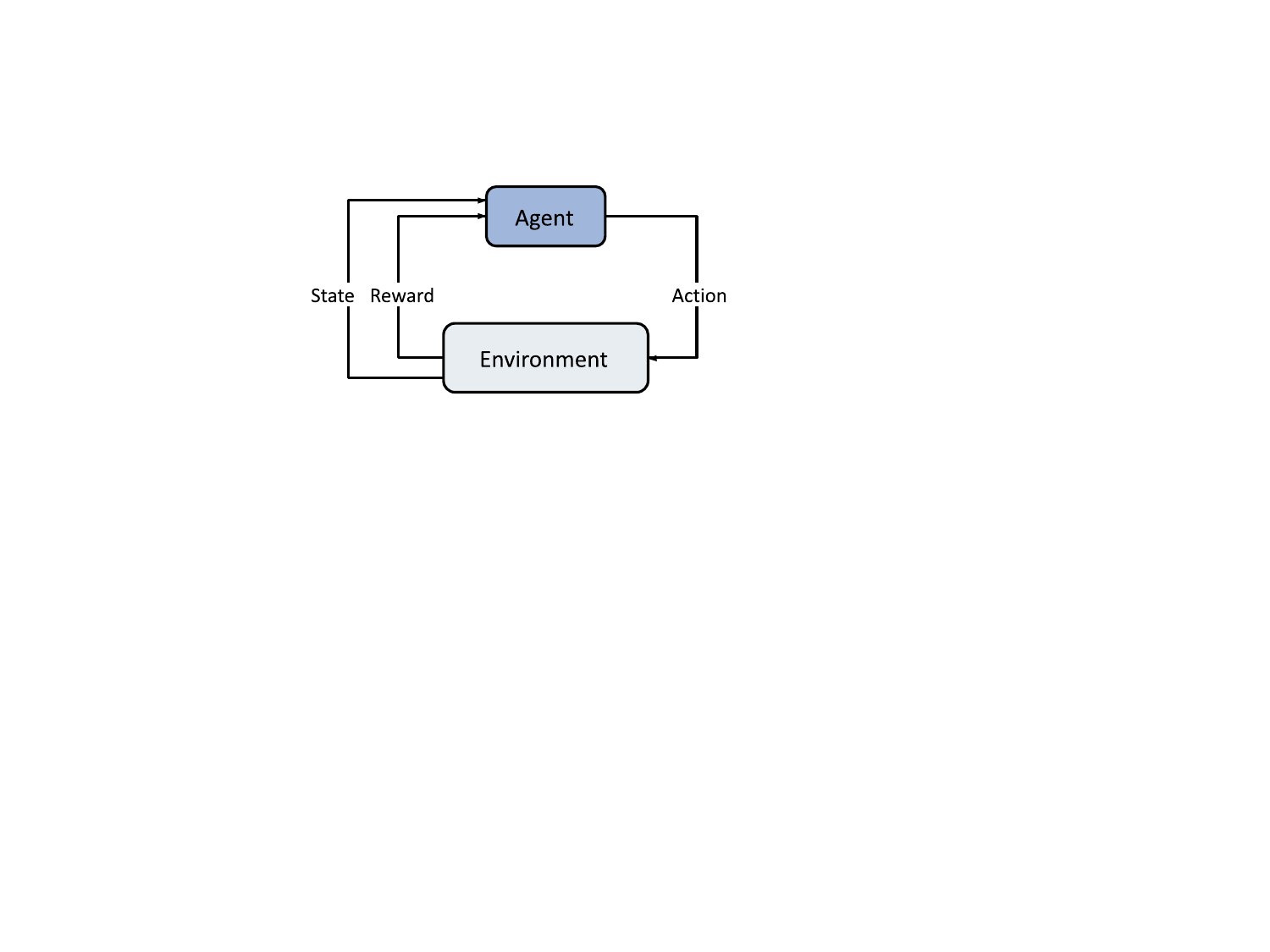} 
\caption{Illustration of Reinforcement Learning process showing how the agent interacts with the environment to maximize the reward.}
\label{fig1}
\end{figure}

\begin{figure*} 
\centering
\includegraphics[width=0.8\textwidth, trim = 2cm 2cm 3cm 4.5cm]{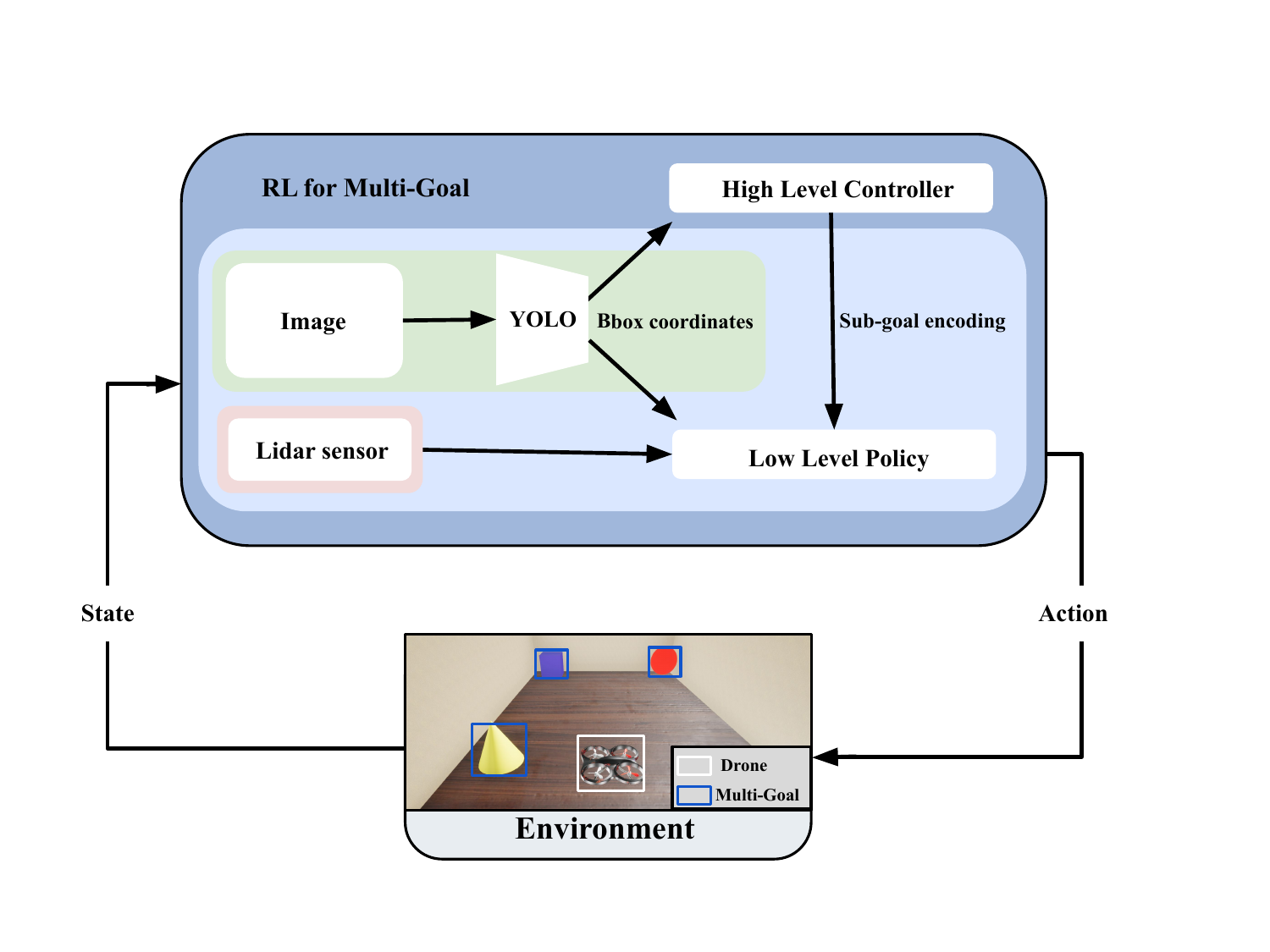} 
\caption{Illustration of the architecture for a hierarchical agent used in this work. The High-level controller which acts as a state machine generates a sub-goal when the agent completes an episode. The Low-level Policy outputs an action at every step and the agent performs the action in the environment to reach the goals.}
\label{fig2}
\end{figure*} 

\section{Related Work}\label{SEC:Rel}

There is considerable work shown in high-fidelity simulators for visual navigation in large-scale indoor settings where control policies are trained for end-to-end navigation. Some methods try to solve the problem of multi-goal detection and navigation for complex tasks in the real world. But there are not many works that use pre-processing on images to solve the problem of Sim2Real for production in the real world. We discuss the related work in the following subsections. 

\subsection{Vision Based Navigation}
Existing research in Vision-Based Navigation tends to pass sensory input into a meaningful state before sending it to a Reinforcement Learning (RL) agent, with the expectation that the robot can implicitly learn the navigation problem through recursive trials.~\cite{Zhu2017TargetdrivenVN} proposed to solve this problem by finding the similarity between different objects using Siamese Network~\cite{Koch2015SiameseNN}.
The work of~\cite{inproceedings} incorporated a meta-learning approach where the agent learns a self-supervised interaction loss during inference to avoid collisions. However, none of these methods use any prior information or semantic context. Moreover, the learned features are expected to vary with the test setting, thereby limiting the scope of this method in unseen environments.

There are a few works on autonomous drone navigation in the real world. In reducing the sim2real gap, the work by ~\cite{ALLAMAA2022385, 9196730, wang2022tacto} shows promising results but requires high-fidelity simulators and high simulation time. But policies trained in the simulated environment do not generalize well in unknown environments. Therefore, purely vision-based navigation seems to lack in production settings.

\subsection{Hierarchical Agent}
Hierarchical goal-based tasks can be modeled as a Markov decision process~(MDP), consisting of a state-space $S$, an action space $A$, a reward function $r$: $S$ × $A$ → $R$, an initial state distribution $s0$, and a transition probability $p(st+1|st, at)$ that defines the probability of transitioning from state $st$ and action $at$ at time $t$ to next state $st+1$. A high-level illustration of the MDP process is shown in Figure~\ref{fig1}.

~\cite{kaelbling1993learning} first proposed an algorithm addressing dynamically changing goals in a multi-goal setting. For each goal object, the algorithm learned a specific value function using Q-learning. More recently,~\cite{pmlr-v37-schaul15} proposed Universal Value Function Approximators~(UVFA), a unique policy able to address an infinite number of goals by concatenating the current state and goal state to feed to the network. This information could be the textual embedding about the target ~\cite{shiri2022efficient}, an image ~\cite{fang2022target}, or the position of the goal in the environment ~\cite{pmlr-v37-schaul15}.

\begin{figure*}

\includegraphics[width=1\textwidth, trim = 0cm 4.5cm 0cm 2.6cm, clip]{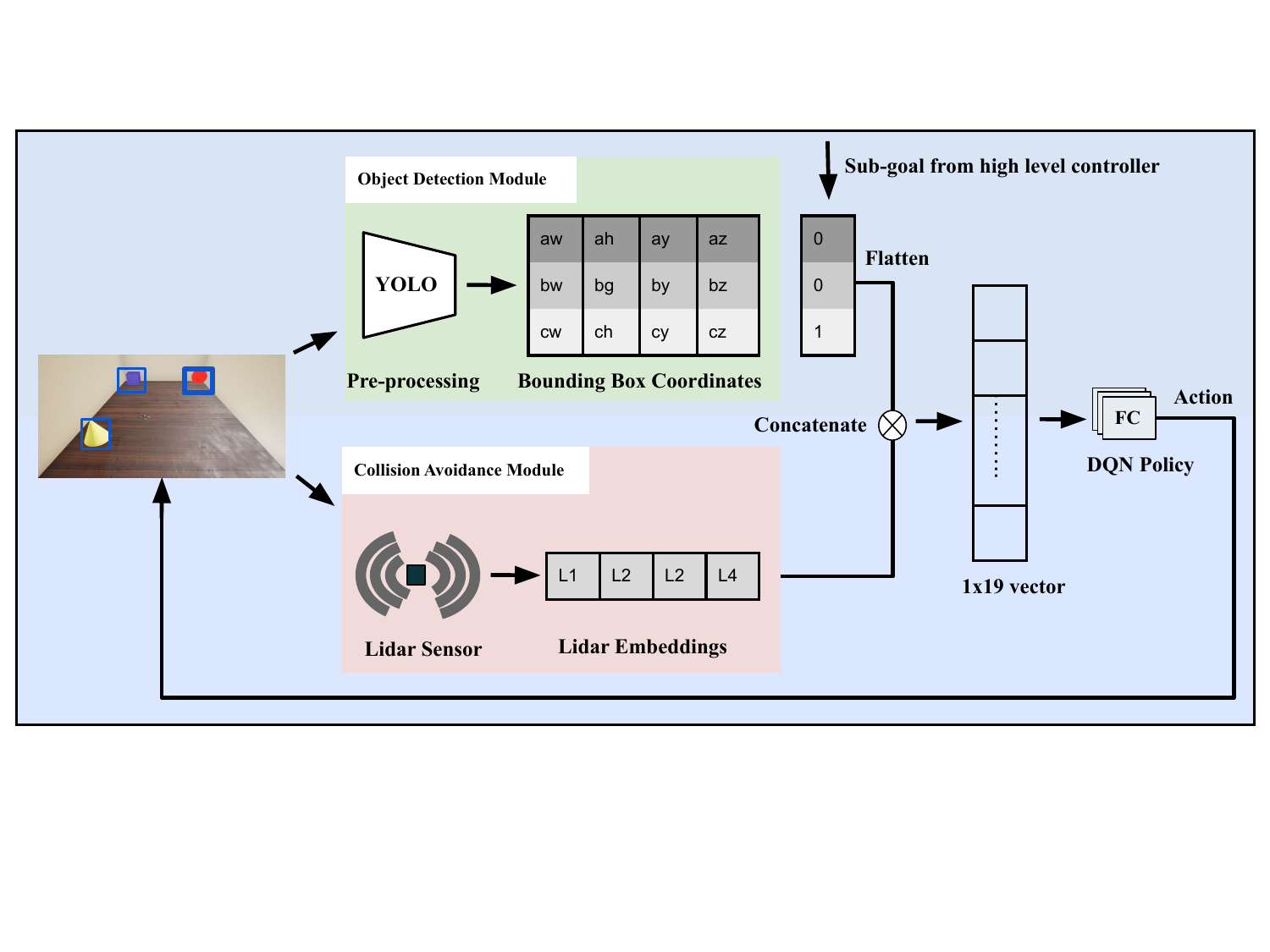} 
\caption{Details of Low-level Policy for goal-dependent navigation. The Bounding box~(Bbox) coordinates from pre-processed image and goal vector (received from the High-level controller shown in Figure~\ref{fig2}) are concatenated with the lidar embedding (used for collision avoidance) as an input to the low-level policy. }

\vspace{-5pt}

\label{fig3}
\end{figure*}

Multi-goal approaches prove better than simply training a policy per goal because knowledge can be transferred between different goals using off-policy learning. A naive solution to the multi-goal problem would be to concatenate the state and goal data as a unique input to the network. But this has proven to not be the best solution as the agent only sees a small subset of mappings between states and goals~\cite{pmlr-v37-schaul15}.  But with reduced state space, such mappings become easier.
With our approach, the state space grows only by a vector of five values for every object and is easily transferable to other environments. Thus, allowing us to use the naive approach of universal Value Function Approximators~(UVFA)~\cite{pmlr-v37-schaul15}.
The overview of the same is elaborated in Figures~\ref{fig2}~and~\ref{fig3}.

\subsection{Object Detection}
Object Detection is used for classifying and localizing objects in a bounding box. A classifier is prepared by training on a set of images in all different methods for object detection. Object detectors are used generally to detect one object out of all the other objects. The models are then evaluated on unseen datasets, and the accuracy is measured accordingly.

There are two types of object detectors - namely, region-based, and region-free methods. Region proposal-based methods include R-CNN, Faster R-CNN~\cite{ren2015faster}, etc., and classification-based methods include YOLO, single shot detector~(SSD)~\cite{liu2016ssd}, etc. 

In~\cite{navardi2022toward}, \Add{our previous work}, we used the DQN agent for single-agent, single-goal navigation. Object detection was performed on grayscale images using a classification-based method - YOLO. In this work, a success rate of 92\% was achieved for navigation towards a single object in simulation. Similar results were observed in a real-world environment. This is shown in Figure \ref{fig6:grayVScolor}.

Our current work outperforms the previous work owing to the colored camera. Most importantly, this work differs from the previous work, as here, the agent learns to navigate towards multiple goals hierarchically, thus showing its generalization towards multi-goal learning. 
While object detectors are mainly used to detect objects, they can also be used for collision avoidance when equipped with the right sensors~\cite{8919366}. But we use four-point lidar embeddings and equivalent distance sensors in the real world to get distances from the obstacles in all four directions (left, right, front, back). 

\begin{figure}
\centering
\includegraphics[width=0.9\columnwidth, trim = 0cm 0cm 0cm 0cm]{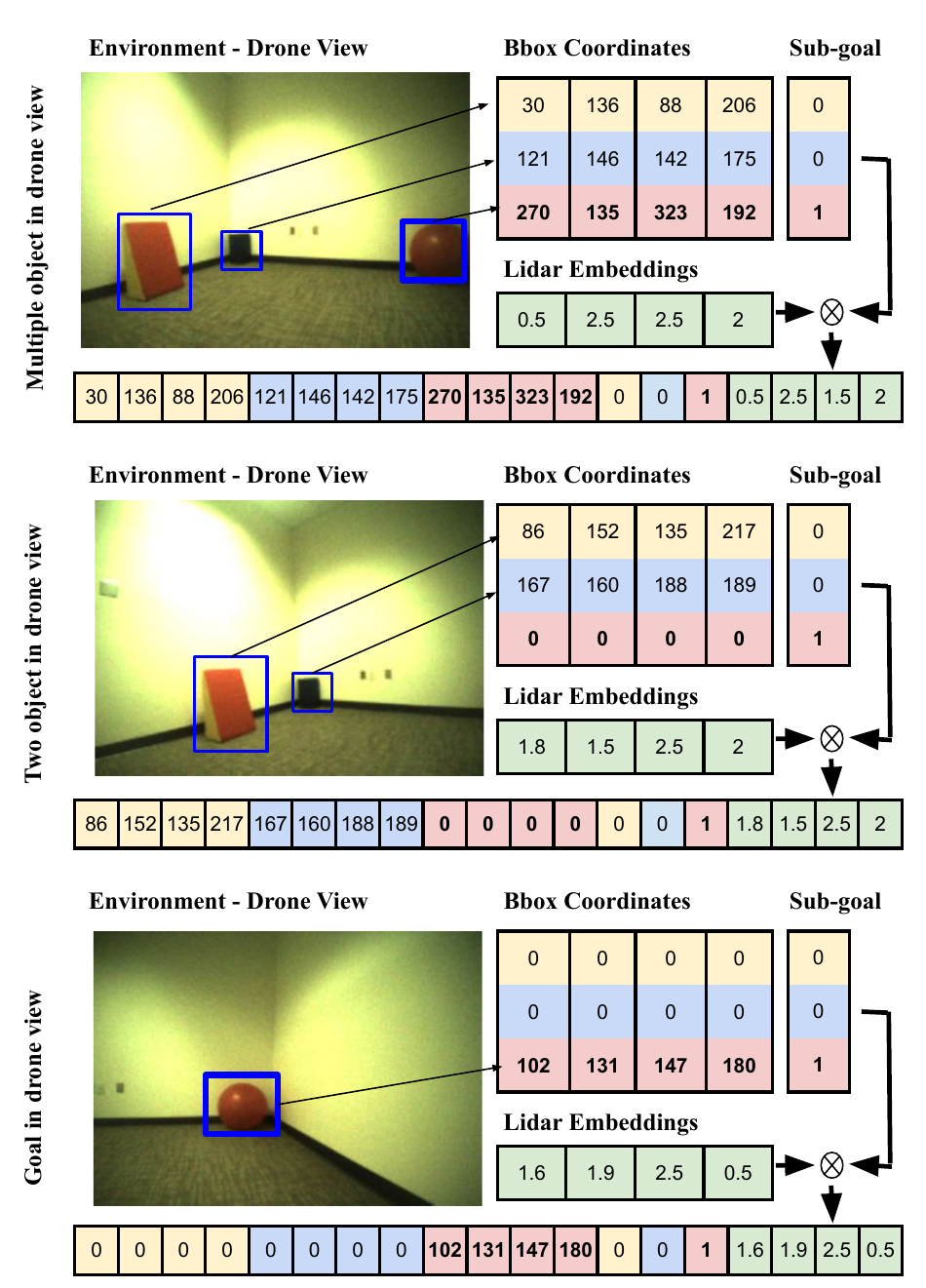} 
\caption{Three different examples 
for state-space showing values of Boundary box~(Bbox) coordinates, goal encoding, and lidar embedding. The Final vector, state space, is the concatenated vector of the three vectors.}
\label{fig4:ex}
\end{figure}

\section{Proposed Methodology}

As illustrated in Figures~\ref{fig2}~and~\ref{fig3}, we design a framework to train hierarchical agents for multi-goal navigation. The proposed architecture consists of a high-level controller that provides sub-goals and a low-level policy that uses pre-processed inputs and lidar readings to perform these sub-goals. The proposed architecture is explained in the following subsections.

\subsection{High-level Controller}

The High-level controller in our method is used purely to give sub-goal instructions to the Low-level policy. Although it can be trained to learn different sequences~\cite{shu2018hierarchical}, we've limited the scope of the High-level controller to provide sub-goals from a fixed sequence and the current observation. 

The High-level controller relies on the current observation to provide sub-goals to the Low-level planner.
The images captured by the agent in the environment are passed through an object detector(YOLOv5) to receive the bounding box coordinates as shown in Figure~\ref{fig2}.  

In the simulation, we switch sub-goals on the successful completion of a sub-task. A sub-task is considered successful when the agent reaches the sub-goal within a distance of 0.1 meters. In a real-world environment, there is no way of knowing the success distance without sensors. One way to address this issue is by setting a minimum number of steps~=~k by when we know the agent can reach the goal. This ``k" can be found by a trial-and-error method in the real world.

\begin{figure*}
\includegraphics[width=1\textwidth, trim = 0cm 5cm 0cm 0cm, clip]{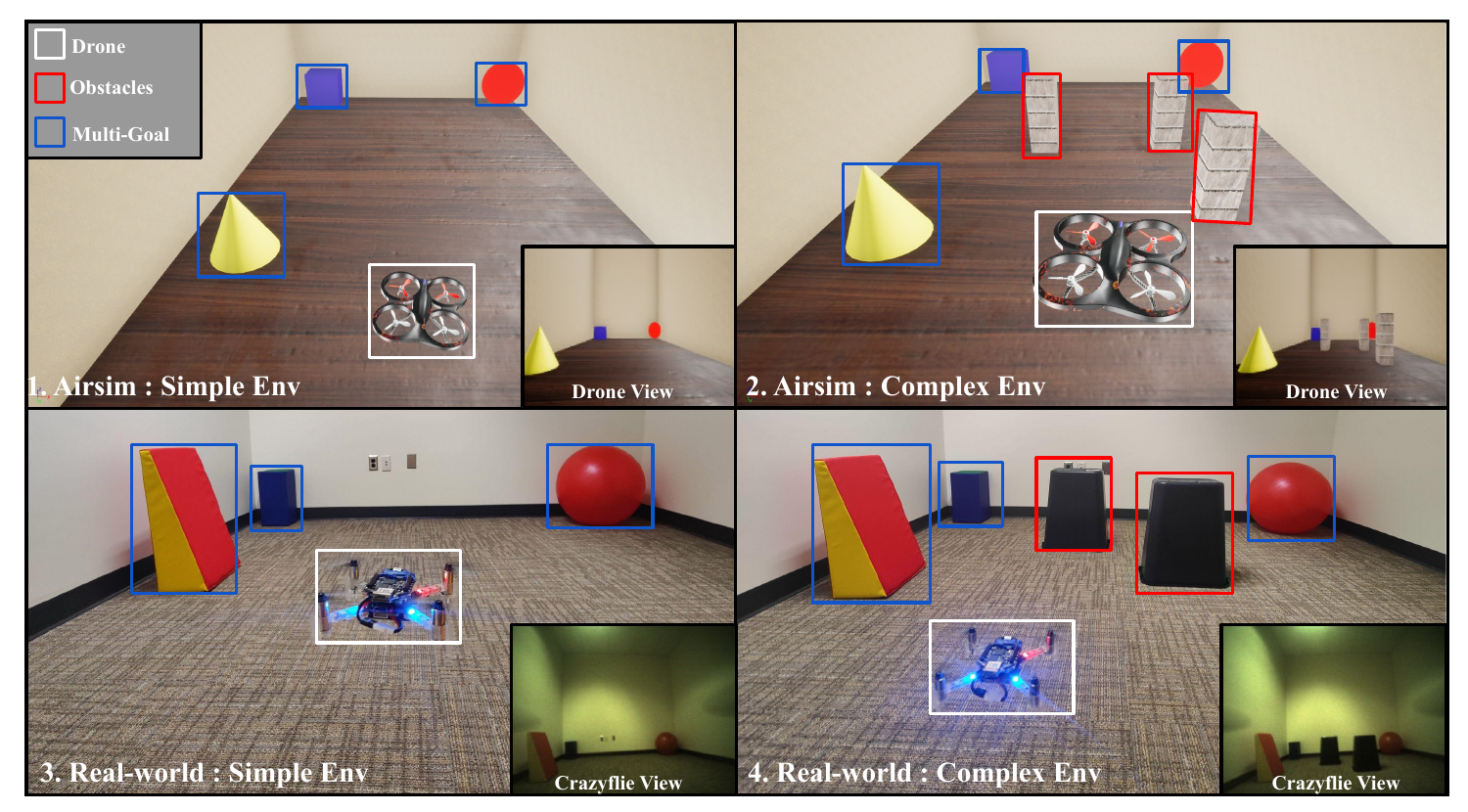} 
\caption{Simple and complex environments for both simulation and real-world environments. The simple environment comprises the multi-goal objects and the UAV. The complex environment has additional obstacles that need to be avoided by the agent while navigating the environment.}
\label{fig5:airsim}
\end{figure*}

A more realistic way of deciding this factor is by defining a precondition based on the size of the image. Since we use bounding box coordinates, we can use the maximum size of the Bounding box(Bbox) as the precondition. When the Bounding box of the goal object reaches a certain size, the High-level-controller can switch the sub-goal. 

The high-level controller uses the latter method to switch sub-goals. Figure~\ref{fig4:ex} describes some examples of Bounding box coordinates. Each Bounding box output consists of 4 predictions: $x1, y1, x2, y2$. The $(x1, y1)$ corresponds to the x and y coordinates of the top left corner of the rectangle. The $(x2, y2)$ corresponds to the x and y coordinates of the bottom right corner of the rectangle. Therefore, $width =(x2 -x1)$ and $height = (y2 - y1)$. In our experiments, the agent is closest to the objects when either the height or width of the bounding box coordinate is at least 70\% of the whole frame's height and weight respectively. In both simulation and real-world environments, the agent operates on constraints of a maximum number of $steps = 50$ to reach the sub-goal. As illustrated in Figures~\ref{fig2}~and~\ref{fig3}, we design a DQN-based deep learning model for the goal-dependent navigation task. The model takes the goal as part of the input and enables the agent to learn a series of different goals jointly. The proposed approach is also used to go to a sequence of goals. 
 
\subsection{Low-level Planner}

The architecture for the Low-level planner is shown in Figure~\ref{fig3} which comprises two modules - an object detection module and a collision avoidance module. The details of the object detection module are described as follows:

First, the pre-processing step consists of a YOLOv5 model. The input image from the environment is processed by the YOLO module to produce outputs in the form of the Bounding box (Bbox) coordinates of the objects. YOLO splits the input image into an $m$x$m$ grid, and each grid generates b bounding boxes and confidence scores for those boxes. 

Second, outputs from the YOLO model in the form of BBox coordinates (1x12) are concatenated with the sub-goal vector(1x3) received from the High-level controller to form a (1x15) vector for three objects.

In the collision avoidance module, the lidar sensor in the simulation environment is used to train the DQN agent to avoid obstacles. This is important because the object detection module only provides details about the goal objects.  This process is represented in Figure~\ref{fig3} in Collision Avoidance Module. The lidar embeddings are then concatenated with the vector from the object detection module to form the state space (1x19).

The final state space is the concatenated vector of the Bounding box coordinates, goal vector, and lidar embeddings, which are inputs to the DQN policy. We use Q-learning as it is more sample-efficient than policy gradient and Monte Carlo value-based methods. The action space is right, left, and straight.

\begin{equation}
R_{t} =
\begin{cases}
R_{goal} \quad d_{t} \le d_p \\      
R_{coll}  \quad d_{o} \le d_q  \\
R_{fail} \quad step = max\_step\\
0
\end{cases}
\end{equation}

We tuned the reward function to learn to navigate to all three objects. The reward function is shown above where $R\textsubscript{goal}$~=~$2~-~0.2~\times~(current\_step/max\_step)$ is the reward on reaching the target position, $R\textsubscript{coll}$~=~$-0.5$ is a negative reward when the agent collides with obstacles in the environment. $R\textsubscript{fail}$~=~$-1.5$ is the penalty when the agent cannot reach the goal object in $max\_step = 300$ and $0$ in other situations. $d\textsubscript{t}$ is the distance between the goal and the drone at time $t$, and $d\textsubscript{o}$ is the distance from the obstacle. $d\textsubscript{p}$ and $d\textsubscript{q}$ are predefined distances used to check if the target is reached or a collision has occurred.

\begin{figure}
     \centering
    \includegraphics[width=7.5cm]{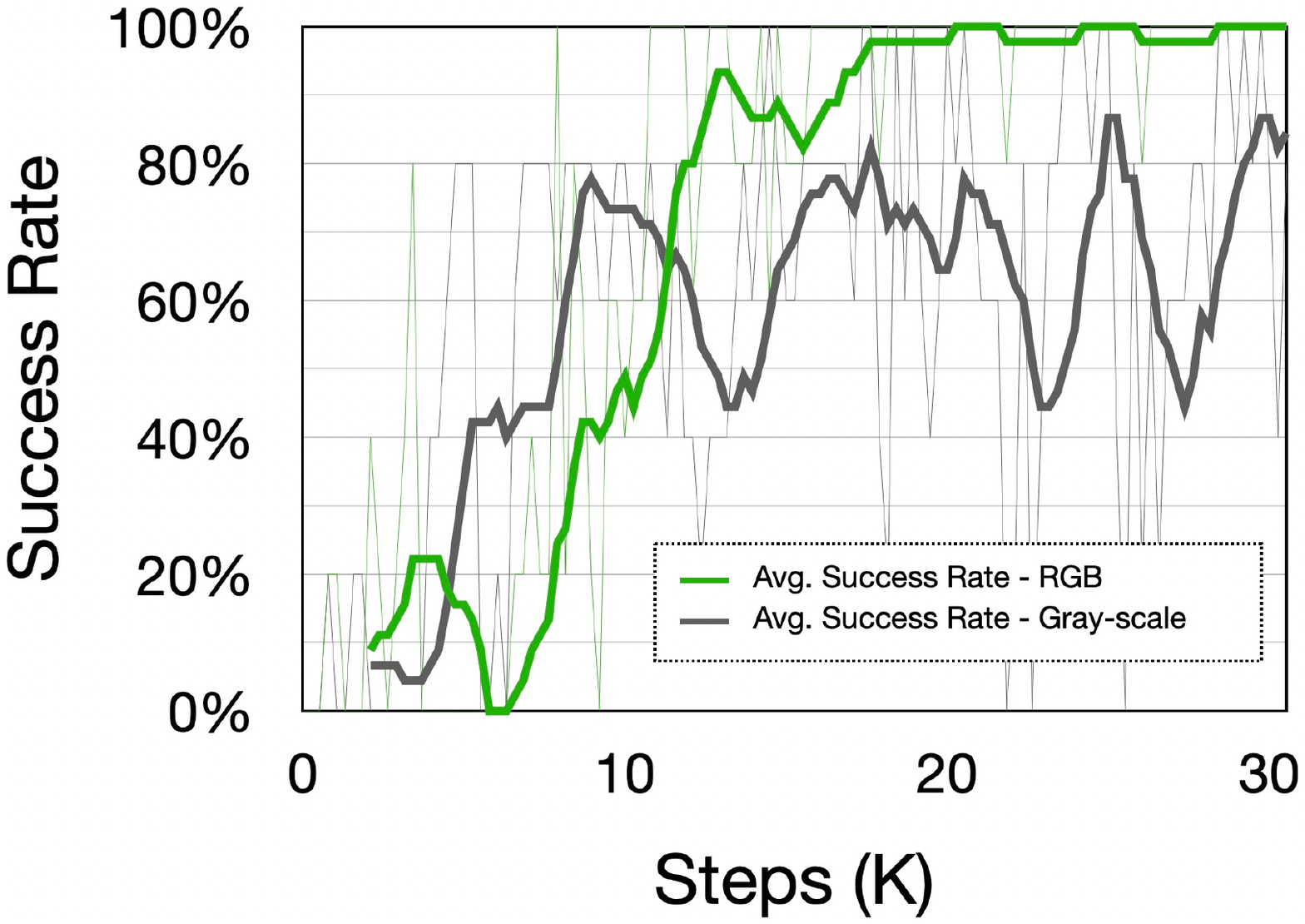}
     \caption{Success rate result of agent training in an environment with one object using grayscale~\cite{navardi2022toward} and RGB camera (ReProHRL). The result illustrates the agent with an RGB camera achieves 100\% success rate in about 20K steps.}
     \label{fig6:grayVScolor}
 \end{figure}

\section{Experimental Setup}
An overview of our simulation environment, training methods, and real-world environment will be discussed in this Section. The performance of the RL agent with the proposed configuration is tested in both simulation and real-world environments with different complexities. The air-learning environment in simple and complex settings is shown in Figure~\ref{fig5:airsim}.

\subsection{Simulation Environment}
In this section, we elaborate on our simulation environment. Our experiments are based on the simulation environment shown in Figure~\ref{fig5:airsim}. Air-Learning~\cite{krishnan2021air}, which is based on Airsim is specifically designed for aerial robotics and autonomous UAVs in 3D environments. It generates high-fidelity photo-realistic environments with domain randomization and flight physics model for the UAVs to fly in. One of the other advantages of Air-Learning environment is that it comes with an energy module that can be used to measure the energy expended by UAVs. This is especially useful when we transfer the models to resource-constrained UAVs since they come with a limited amount of battery. We designed an environment with a 25x15m room and multiple goals in it. In addition to the environment with only goal objects, we also evaluate the performance of the proposed methodology in a complex environment shown in Figure~\ref{fig5:airsim} (2).

\begin{figure*}
\includegraphics[width=\textwidth]{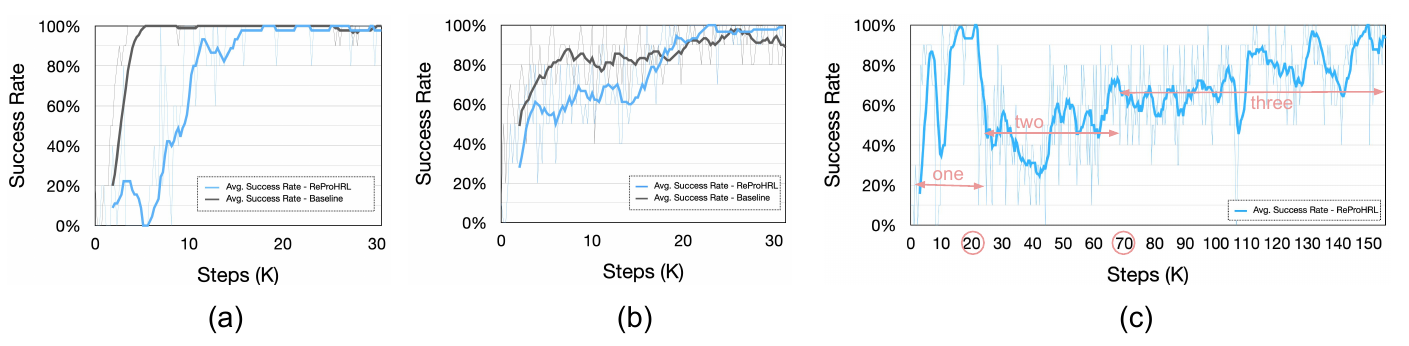} 
\caption{Success rate achieved by the proposed approach (ReProHRL) in the Air-Learning environments with respect to the number of episode steps. For the baseline, we used the language-guided approach presented in~\cite{shiri2022efficient}. We reported results for simple environments with (a) one object, (b) two objects, and (c) three objects. For three objects shown in (c), the DQN agent trained using curriculum learning shows training for one object for 20K steps, two objects for 70K steps, and finally for three objects until 150K steps.}
\label{fig7:SR}
\end{figure*}

\begin{figure}
     \centering
    \includegraphics[width=7.5cm]{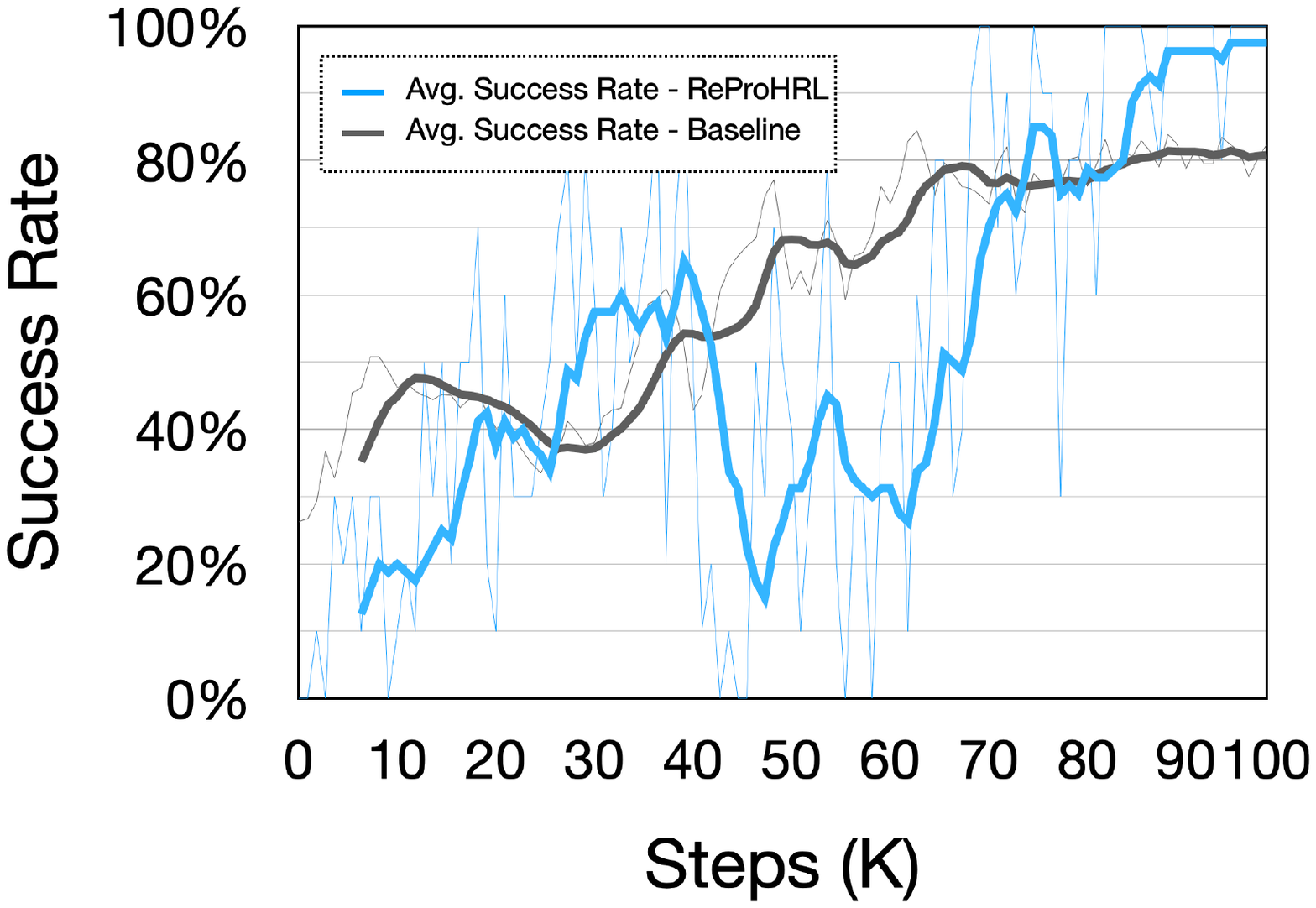}
     \caption{Success rate graph for agent trained to navigate to a single-goal in a complex environment with multiple obstacles that is shown in Figure~\ref{fig5:airsim} (2).}
     \label{fig8:cmplx}
 \end{figure}

\subsection{Training Details}
We first trained YOLOv5 on the training data sampled from both Air-learning and real-world environments. We sampled a total of 4k images. Although the pre-trained models can be used for real-world object detection, to improve the accuracy and performance of our model, we manually annotated the images and trained them on the YOLOv5 model. These images are trained with YOLOv5s to fit the low-powered drone. The final map score for YOLOv5s is 0.96 which is over 0.05 more than when trained with Gray-scale images. 

\subsection{Non-Curriculum Learning}
We train the DQN agent to navigate to one or two objects in both simple and complex environments shown in Figure~\ref{fig5:airsim} (1) and (2) respectively. To determine the performance of our algorithm compared to the baseline, we train each agent until we reach a satisfactory success rate using non-curriculum learning. We randomize the position of the goal every episode to be anywhere in the arena. The checkpoints are saved every 10000 steps and we use the last saved checkpoint for evaluation and further experiments.

\subsection{Curriculum Learning} 
To improve the performance of the DQN agent in a multi-goal setting, we use a curriculum learning methodology where the task is progressively made difficult - the agent first learns to navigate to a single object and then to two objects and finally three. As seen in Figure~\ref{fig7:SR} (c), the success rate curve goes up quickly before dropping down since the goal in the initial steps was to navigate to a single object.

\subsection{Real-world Deployment}
The Crazyflie platform features easily replaceable parts, so we use Crazyflie to bridge the ‘Sim2Real’ gap and to test directly in our real-world experiments. Crazyflie weighs 27 grams and has dimensions 92 mm×92 mm×29 mm. Stackable chips like AIdeck, a Multi-ranger deck, a location module, etc., can be used for additional processing, location, and other functionalities. The processing is done on an ARM CortexM4 operating at 160 MHz on an STM32F405 microcontroller, with 192 KB of SRAM and 1024 KB of flash memory. 
We use the same Yolo model which was trained on different environments for our real-world evaluation. 

Deployment of models on UAVs is a challenging task especially on resource-constrained UAVs since they need to accomplish the given task with limited energy. Additionally, onboard memory is scarce~\cite{shiri2022efficient}, and RL models are computationally intensive. Therefore, we need to design the models carefully. As the size of the drone decreases, its battery capacity also decreases, which aggravates the problems. Therefore, we carefully monitor the energy both in simulation and real-world environments. We give a higher penalty to the agent if it fails to complete the task to encourage faster task completion. On average, the agent completes the task under 25 steps in a 15mx25m room operated under actions of lower time duration.

In the real world, to avoid collisions, we used a multi-ranger deck, an extension deck for Crazyflie. This gives the distance from the obstacle to the drone in all four directions, similar to our lidar readings. In the real world, we performed experiments in a 5mx5m room similar to our simulation environment as shown in Figure~\ref{fig5:airsim} (3) and (4), simple and complex environments, respectively. The actions and action duration of Crazyflie is tuned for better efficiency in the real world.

\section{Experimental Results}
This section discusses the evaluation of the proposed method on the Air-learning environment, an open-source setting within the Unreal game engine that enables navigation in near-photo-realistic indoor scenes and real-world settings. The performance is evaluated quantitatively in terms of success rate - the number of times the agent reaches the goal over the total number of episodes. For each episode, we randomly initialized the starting location of the agent in the same simulation environment with a different number of goals, i.e., one, two, and three in a simple and complex environment. 

\begin{table*}[t]
\centering
\begin{tabular}{|c|l|llll|}
\hline
\multirow{4}{*}{Methods}  & \multicolumn{1}{c|}{\multirow{4}{*}{Number of Goals}} & \multicolumn{4}{c|}{Environment}                                                                                                                                                                                                                                                                                                                                                                                            \\ \cline{3-6} 
                          & \multicolumn{1}{c|}{}                                 & \multicolumn{2}{c|}{Simple Env}                                                                                                                                                                                         & \multicolumn{2}{c|}{Complex Env}                                                                                                                                                                  \\ \cline{3-6} 
                          & \multicolumn{1}{c|}{}                                 & \multicolumn{1}{|l|}{\multirow{2}{*}{\begin{tabular}[c]{@{}l@{}}Success Rate\\         (\%)\end{tabular}}} & \multicolumn{1}{l|}{\multirow{2}{*}{\begin{tabular}[c]{@{}l@{}}Training time\\       (steps)\end{tabular}}} & \multicolumn{1}{l|}{\multirow{2}{*}{\begin{tabular}[c]{@{}l@{}}Success Rate\\        (\%)\end{tabular}}} & \multirow{2}{*}{\begin{tabular}[c]{@{}l@{}}Training Time\\       (steps)\end{tabular}} \\
                          & \multicolumn{1}{c|}{}                                 & \multicolumn{1}{l|}{}                                                                                     & \multicolumn{1}{l|}{}                                                                                       & \multicolumn{1}{l|}{}                                                                                    &                                                                                        \\ \hline
\multirow{3}{*}{Baseline} & Single goal                                           & \multicolumn{1}{l|}{100\%}                                                                                & \multicolumn{1}{l|}{7K}                                                                                     & \multicolumn{1}{l|}{80\%}                                                                                & 90K                                                                                    \\ \cline{2-6} 
                          & Two goals                                             & \multicolumn{1}{l|}{95\%}                                                                                 & \multicolumn{1}{l|}{25K}                                                                                    & \multicolumn{1}{l|}{-}                                                                                   & -                                                                                      \\ \cline{2-6} 
                          & Three goals                                           & \multicolumn{1}{l|}{-}                                                                                    & \multicolumn{1}{l|}{-}                                                                                      & \multicolumn{1}{l|}{-}                                                                                   & -                                                                                      \\ \hline
\multirow{3}{*}{ReProHRL} & Single goal                                           & \multicolumn{1}{l|}{100\%}                                                                                & \multicolumn{1}{l|}{15K}                                                                                    & \multicolumn{1}{l|}{98\%}                                                                                & 100K                                                                                   \\ \cline{2-6} 
                          & Two goals                                             & \multicolumn{1}{l|}{100\%}                                                                                & \multicolumn{1}{l|}{22K}                                                                                    & \multicolumn{1}{l|}{-}                                                                                   & -                                                                                      \\ \cline{2-6} 
                          & Three goals                                           & \multicolumn{1}{l|}{\Add{95\%}}                                                                                & \multicolumn{1}{l|}{150K}                                                                                   & \multicolumn{1}{l|}{-}                                                                                   & -                                                                                      \\ \hline
\end{tabular}

\caption{Summary of Success rate and Training time comparison between our approach and the baseline\Add{~\cite{shiri2022efficient}} all trained for under 150K steps in Air-learning Environment shows that our approach outperforms the baseline while taking fewer steps in multi-goal settings.}
\label{tab1:success}
\end{table*}

\subsection{Simulation Results}

To see how our method compares to state-of-the-art baseline~\cite{shiri2022efficient}, an ablation study is made between the two approaches for a different number of goals, i.e., one, two goals, and three goals in both simple and complex environments. The baseline is a visual navigation-based model trained in simple and complex environments. \Add{The baseline~\cite{shiri2022efficient}, one of our previous works encodes the goal information that is in the form of English language instructions into a vector. Therefore, state space comprises of a goal-state encoding similar to our approach.}

The learning curves in Figure~\ref{fig7:SR} show that our approach obtains better performance than state-of-the-art in most cases, both in terms of convergence speed and success rate. For single object navigation, both models performed exceptionally due to the simpler task.In navigating towards two goal objects that are selected at random, our approach reached a convergence to near 100\% success rate at 25k steps, and the baseline failed to reach 100\% even after 1 million steps. This shows that our approach performs better in multi-goal settings even with a pre-processing step.
Finally, for the three-object navigation setting, our method reached a 95\% success rate in 150k steps. In the complex setting shown in Figure \ref{fig5:airsim} (2), with our approach, the success rate reaches 98\% at 100k steps but the stat-of-the-art baseline reaches only 80\% as shown in Figure \ref{fig8:cmplx}.

The performance could be attributed to our reduced state space that can capture the navigation-relevant association between state and goal and encourage cooperation when training an agent to navigate towards multiple goals. 

For single and two-goal navigation settings, we trained the DQN agent using a non-curriculum approach. But, to improve the training time for navigation in a three-goal setting, we used a curriculum learning approach where we trained for one object until the DQN agent learned to navigate to a single object in 20K steps. Next, we trained the agent for two goals until 70K steps. It learned quickly for single-object navigation, but the success rate dropped low when the complexity increased before the agent learned to navigate to all three objects at 1.5 million steps. The success rate graph in Figure~\ref{fig7:SR}~(c) reflects the same. 

All models were trained to reach an acceptable success rate. But any training above 1.5 million steps is time-consuming in high-fidelity simulators. The Success Rate for all methods is summarized in Table \ref{tab1:success}.
 
\subsection{Performance in Real-world} 

In ~\cite{navardi2022toward}, a single object-based navigation model trained on grayscale images in the simulation was deployed in the real-world environment. It also shows that the drone reaches the goal as long as it is within a visual distance from the object. With ReProHRL, the performance of the YOLO model is better on account of RGB images used in training. Additionally, the model generalizes well to unseen observations, which is vital for the production of RL algorithms.

Though we observe an improvement in single-goal-based navigation, both in terms of performance and time, the onboard energy proves insufficient for multi-goal navigation as the drone takes around 30 seconds to 1 minute in a 5mx5m room to reach an object. Although the battery lasts up to 7 minutes on Crazyflie, with the addition of 2 decks and additional processing, the energy lasts at most for 3 minutes. In high-performance UAVs like DJI-Mavic Pro ~\cite{elkhrachy2021accuracy} that have 15 times more onboard energy than Crazyflie, this would not be concerning.

\section{Conclusion}

In this work, we proposed a Ready for Production Hierarchical Reinforcement Learning~(ReProHRL) agent scheme that enables the agent to learn the skill of navigating to multiple goals hierarchically which can be deployed in the real world. We introduced two components in our architecture: a High-level controller and a Low-level planner to enhance sequential goal completion in visual navigation. We validated the proposed approach by experimenting with navigation tasks for multi-goal navigation and comparing it with two state-of-art works. Experimental results in the Air-learning environment demonstrated the superiority of the proposed architecture, ReProHRL, over existing methods in goal-dependent navigation tasks. With the proposed approach, we achieved near 100\% success rate in the simple environment, multi-goal setting as well as the complex environment,  single-goal setting. The latter has  18\% improvement over the baseline~\cite{shiri2022efficient}. These settings resulted in a fast response time and adaptation to unseen environments. Moreover, we have evaluated the proposed approach in physical environments by deploying it on a drone named Crazyflie in a real room with multi objects.

\section{Acknowledgement}
This project was sponsored by the U.S. Army Research Laboratory under Cooperative Agreement Number W911NF2120076.

\nobibliography{aaai22, eehpc}

\bigskip

\section{References}

\bibentry{zuluaga2018deep}

\bibentry{duisterhof2021tiny}

\bibentry{navardi2022optimization}

\bibentry{truong2021bi}

\bibentry{zhang2021sim2real}

\bibentry{duisterhof2021sniffy}

\bibentry{zhang2020sim2real}


\bibentry{nguyen2019reinforcement}



\bibentry{benjumea2021yolo}

\bibentry{doersch2019sim2real}

\bibentry{taghibakhshi2021local}

\bibentry{prakash2021interactive}

\bibentry{staroverov2020real}

\bibentry{li2020hrl4in}


\bibentry{pmlr-v37-schaul15}




\bibentry{wang2022tacto}

\bibentry{kaelbling1993learning}

\bibentry{navardi2022toward}

\bibentry{shiri2022efficient}

\bibentry{giernacki2017crazyflie}

\bibentry{Zhu2017TargetdrivenVN}.

\bibentry{Koch2015SiameseNN}.

\bibentry{inproceedings}.

\bibentry{ALLAMAA2022385}


\bibentry{9196730}

\bibentry{fang2022target}

\bibentry{ren2015faster}

\bibentry{liu2016ssd}

\bibentry{8919366}


\bibentry{shu2018hierarchical}

\bibentry{krishnan2021air}

\bibentry{elkhrachy2021accuracy}

\bibentry{prakash2021semantic}

\bibentry{prakashhierarchical}

\end{document}